\documentclass[11pt,a4paper]{article}
\usepackage[hyperref]{emnlp-ijcnlp-2019}
\usepackage{hyperref}
\usepackage{times}
\usepackage{latexsym}
\usepackage{graphicx}
\usepackage{url}

\aclfinalcopy 

\title{reproducing ``ner and pos when nothing is capitalized"}

\author{Andreas Kuster \\
    {\tt kustera@ethz.ch} \\\And
    Jakub Filipek \\
    {\tt balbok@uw.edu} \\\And
    Viswa Virinchi Muppirala \\
    {\tt virinchi@uw.edu}}
\date{}

\makeatletter
\def\endthebibliography{%
	\def\@noitemerr{\@latex@warning{Empty `thebibliography' environment}}%
	\endlist
}
\makeatother

\begin{document}
\maketitle
\begin{abstract}
    Capitalization is an important feature in many NLP tasks such as Named Entity Recognition (NER) or Part of Speech Tagging (POS). We are trying to reproduce results of paper which shows how to mitigate a significant performance drop when casing is mismatched between training and testing data. In particular we show that lowercasing 50\% of the dataset provides the best performance, matching the claims of the original paper. We also show that we got slightly lower performance in almost all experiments we have tried to reproduce, suggesting that there might be some hidden factors impacting our performance. Lastly, we make all of our work available in a \href{https://github.com/andreaskuster/uw-nlp}{public github repository}.
\end{abstract}

\section{Introduction}
Previous works have shown that there is a significant performance drop when applying models trained on cased data to uncased data and vice-versa \cite{wang-etal-2006-capitalizing}. Since capitalization is not always available due to real world constraints, there have been some methods trying to use casing prediction (called \textit{truecasing}) to tackle this trade-off.

The work we reproduce tries to battle this issue in two popular NLP tasks by lowercasing 50\% of the original datasets.

\section{Contributions}
This paper effectively shows how well work from \cite{ner-and-pos-original} can be reproducede and how well it applies to a few other settings.\\
The original paper does show how casing issue in NLP can be effectively solved through a method which requires close to none overhead in terms of development time, and no additional overhead in runtime, especially when compared to methods such as truecasing (This is true for mixing 50\% cased and 50\% uncased, if we mix full datasets of both then training will take approximately twice as long).\\
It also serves as a reproducing work to show that truecasing performance can be reproduced.

\subsection{Hypotheses from original paper}
\label{sec:original_hypotheses}
Original paper proposes following hypotheses:
\begin{itemize}
    \item Truecasing fits strongly to data it is presented leading to performance drops on different datasets.
    \item Mixing cased and uncased data provides the best performance in NER task on CoNLL 2003 dataset. This is not due to larger size of the dataset, since the same method works if we only lowercase 50\% of the original dataset.
    \item Such techinque generalizes well on noisy datasets such as Twitter data.
    \item Mixing cased and uncased data provides the best performance in POS task on Penn Treebank dataset. This is not due to larger size of the dataset, since the same method works if we only lowercase 50\% of the original dataset.
\end{itemize}

\subsection{Hypotheses addressed in this work}
\label{sec:current_hypotheses}
In addition to hypotheses tested in Section~\ref{sec:original_hypotheses} we also tested:
\begin{itemize}
    \item POS:
    \begin{itemize}
        \item Mixing both cased and uncased data leads to the best performance regardless of word \textbf{embedding} used.
        \item Mixing both cased and uncased data leads to the best performance regardless of word \textbf{dataset} used.
        \item Using ELMo model with CRF layer in POS task to outperform the one without, regardless of data casing technique. This is based on \cite{BiLSTM-CRF}.
    \end{itemize}
\end{itemize}

\subsection{Experiments}
We conducted three main experiments from the paper, which were done roughly in parallel to each other.

Truecasing is a task where given all lowercase sentences we have to predict correct casing. Here we trained a simple Bidirectional LSTM with a logical layer on top, as described in both \cite{ner-and-pos-original} and \cite{susanto-etal-2016-learning}. Since the former paper, does a great job of mentioning hyper-parameters used in the network, the hyperparameter search is not required.

Part of Speech tagging is a task in which given a sentence we have to predict part of speech for each word. It is our the most in depth experiment, which lead a lot of additional hypotheses which got tested. Firstly, we had to find optimal hyper-parameter setting. The search was not too large, since a few of the parameters (such as number of layers) were known, but it still took a significant amount of time. After finding such setting we compared its results with ones reported the original paper. Then we investigated whether claims from this paper are applicable to other datasets and encodings. Lastly, we looked at the CRF layer, and what its impact on performance on the whole model is.

Named Entity Recognition is a task in which given a sentence we have to predict which parts of it describe some entity. It had the most sophisticated model out of all the experiments we tried reproducing. Fortunately we did not have to do a hyperparameter search since \cite{DBLP:journals/corr/LampleBSKD16} provided a similar model to that of the original paper.

\section{Code}
There are no public repos which do mention the project in their \textit{READMEs}.

However, in case of truecasing, primary author of \cite{ner-and-pos-original} has two repositories: \href{https://github.com/mayhewsw/truecaser}{truecaser} and \href{https://github.com/mayhewsw/pytorch-truecaser}{python-truecaser}. The former one refers to the original implementation from \cite{susanto-etal-2016-learning}, while the python one is a port to python. The (probably) primary author of \cite{susanto-etal-2016-learning} also has a \href{https://github.com/raymondhs/char-rnn-truecase}{github repository} possibly related to truecaser we are trying to reproduce. However, to validate results and test ease of application we decided to not use either of these resources and focus on custom implementation. This also removes dependence on Andrej Karpathy's \href{https://github.com/karpathy/char-rnn}{char-rnn} all of the above mentioned repositories are forks of.

We were not able to find publicly available code for either NER or POS parts of the original paper.

Hence we needed to reimplement the code from scratch. All our work is in \href{https://github.com/andreaskuster/uw-nlp}{public github repository}. We tried to make all results as easily accessible as possible, which means that there is significant overlap between this report and the \textit{README} on that repository.

Truecasing and NER were implemented using PyTorch \cite{pytorch}, while POS was implemented in Keras \cite{keras}, on top of TensorFlow \cite{tensorflow2015-whitepaper}.

\section{Experimental setup and results}

\subsection{Datasets}
\label{sec:datasets}
Datasets were a common resource about all three parts of experiments. Hence we will describe them separate here, and for each experiment specify which exact dataset was used:
\begin{itemize}
    \item CoNLL2003 - From \cite{conll2003}. Not publicly available. Shared as part of class.
    \item Peen Tree Bank (PTB) - From \cite{penn-treebank}. Not publicly available. Shared as part of class. We used a loader from \href{https://www.nltk.org/}{nltk} (after appending nonpublic data to it). In particular:
        \begin{itemize}
            \item Sections 0-18 are used for training
            \item Sections 19-21 are used for development, i.e. choosing hyperparameters
            \item Sections 22-24 are used for test, i.e. reporting accuracy
        \end{itemize}
    \item Twitter - From \cite{twitter-dataset}. Specifically we found \href{https://github.com/GateNLP/broad_twitter_corpus}{authors github repository} with data available.
    \item Wikipedia - From \href{https://github.com/raymondhs/char-rnn-truecase/tree/master/data/wiki}{github repository}, which we suspect is authored by primary author of \cite{susanto-etal-2016-learning}. This data was used for truecasing experiment.
\end{itemize}

Additionally, in case of NER and POS each of these datasets occurs in 5 different experimental flavors:
\begin{itemize}
    \item Cased (C) - Standard dataset, as downloaded.
    \item Uncased (U) - Uncased dataset. It's a standard dataset, with an additional step of converting everything to lowercase.
    \item Cased + Uncased (C + U) - Combination of both cased and uncased flavors. Hence it is twice the size of either cased or uncased.
    \item Half Mixed (C + U 50) - Combination of cased and uncased flavors. However, only 50\% of data is lowercase, and remaining half is as-is.
    \item Truecase Test (TT) - Truecased Test dataset. Training data is the same as cased one. Test is converted to uncased version, and then a truecaser is run on it. The Truecaser is not trained on any data, and is provided (in case of this paper, truecaser trained on Wikipedia dataset is used).
    \item Truecase All (TA) - Truecased Train and Test dataset. The transformation described for test in TT is applied to both train and test. The Truecaser is not trained on any data, and is provided (in case of this paper, truecaser trained on Wikipedia dataset is used).
\end{itemize}

\subsection{Padding}
To remove repetition in description of experiments we also want to discuss padding and how results should be understood.

All experiments described below used padding to the maximum sentence size for training. However, for results on both validation (development) and test such padding was either removed or ignored, and hence it is not affecting the results reported in this paper.

\subsection{Truecasing}
\label{sec:exp-truecase}

    \subsubsection{Model description}
    We used the exact same model as described in \cite{susanto-etal-2016-learning}, a 2 layer, bidirectional LSTM. Since encoding was not specified, nor any character level encoding was mentioned in class we used a PyTorch \cite{pytorch} Encoder layer, which learned weights from bag-of-words to a feature-sized word representation. Note that in this model size of word embeddings is equal to hidden state size (both 300 dimensional). Then on top of these two layer a binary linear layer was applied.

    Hence output of this model is 2 dimensional, one specifying that character should be upper cased (true), other that character should be left as is.

    Implementation of this model took 2 hours, mostly due to transposes required by LSTM layers in PyTorch framework. It also helped that assignment 2 required us to implement LSTM model, and we could reuse parts of it.

    \subsubsection{Hyperparameters}
    We used a 2 layer, bidirectional LSTM with hidden size of 300. On top of that a linear layer, with 2 classes was applied, resulting in binary output. We used Adam \cite{adam} optimizer with default settings (learning rate of 0.001, betas of 0.9 and 0.999), and batch size of 100, as suggested in the \cite{susanto-etal-2016-learning}. Since these were the hyperparameters specified, we did not perform hyperparameter search.

    We also used an Out-Of-Vocabulary (OOV) rate of 0.5\%. Due to initial mistake we first defined out-of-vocabulary rate in following manner. When each token was drawn it had a 0.5\% chance to be masked with OOV token.

    However, after clarification from TA we switched OOV to be done at the initialization of tokens. In particular sorted tokens in order of increasing occurrence, and masked all least occurring tokens (such that their sum contained at least 0.5\% of total amount of tokens) with OOV token. This skewed our OOV distribution towards uncommon tokens. This is a desired behavior, due to the fact that the differences between datasets will be in these rare-occurring tokens.

    For wikipedia dataset, with first approach we got a dictionary of size 64, and with second approach it was 41.

    \subsubsection{Results}
    We trained these models for 30 epochs, with crossentropy loss. We recorded both training and validation loss for both, and chose the models with lowest validation loss for each of OOV settings. In both cases, such point was reached after around 7-8 epochs.

    The performance of the model matched one claimed both by original paper, and original truecasing paper on Wikipedia dataset. However, there was a significant mismatch in performance on other datasets provided in the original paper. This can be seen in Table \ref{tab:truecase-result}. A drop-off between datasets is expected, because as explained in original paper, various sources will use various casing schemes specific to the domain.
    \begin{table}[h]
        \centering
        \begin{tabular}{|l|c|c|}
            \hline
            Dataset & Initial OOV & Proper OOV \\
            \hline
            Wikipedia & 92.65 & 92.71\\
            \hline
            \hline
            CoNLL Train & 66.03 & 65.32 \\
            \hline
            CoNLL Test & 63.49 & 63.28 \\
            \hline
            PTB 01-18 & 78.53 & 78.73 \\
            \hline
            PTB 22-24 & 78.47 & 78.69 \\
            \hline
        \end{tabular}
        \caption{F1 scores of best performing model (chosen on validation set of Wikipedia dataset) for various datasets. Initial OOV stands for at-the-read techinque of creating OOV tokens, while proper is the former one of two described above.}
        \label{tab:truecase-result}
    \end{table}
    For reference Table \ref{tab:trucase-original-result} contains results from investigated papers. We can see that there is not a high difference in performance between results for Wikipedia dataset, but there is a 10\% difference in both CoNLL and PTB between our results and original ones. The possible reason for it might be the fact that original paper used fork of, before mentioned, \href{https://github.com/karpathy/char-rnn}{char-rnn}. We looked at it, searching for differences between our intuition and actual implementation, but could not find any.
    \begin{table}[h]
        \centering
        \begin{tabular}{|l|c|}
            \hline
            Dataset & Original \\
            \hline
            Wikipedia (Susanto) & 93.19 \\
            \hline
            Wikipedia & 93.01 \\
            \hline
            \hline
            CoNLL Train & 78.85 \\
            \hline
            CoNLL Test & 77.35 \\
            \hline
            PTB 01-18 & 86.91 \\
            \hline
            PTB 22-24 & 86.22 \\
            \hline
        \end{tabular}
        \caption{F1 scores from reference papers. All entries but ``Wikipedia (Susanto)" are from \cite{ner-and-pos-original}.}
        \label{tab:trucase-original-result}
    \end{table}
    This leaves us in conclusion that, in terms of performance, this experiment can be only partially reproduced, and it is possible that additional techniques (such as dropout, etc.) were applied during training of the model. However, in terms of between-datasets trends the paper describes, we have confirmed them, and \textbf{were able to (partially) reproduce the first hypothesis of the paper}.

\subsection{Part of Speech Tagging}
\label{sec:exp-pos}

    \subsubsection{Model description}
    Model for POS task is not well described in the original paper, and rather references other papers for implementation details \cite{ma-hovy-2016-end}. This left us with only a rough sketch of a model, and a hyperparameter search for most of variables.

    Model used here is a single BiLSTM layer, followed by a Time Distributed Dense layer, followed by an output CRF layer. We used ELMo as the encoding for the input to BiLSTM layer.

    \subsubsection{Hyperparameters}
    \label{sec:exp-pos-hyper}
    As mentioned before, the task required a hyperparameter search. It was performed on standard (Cased) version of PTB dataset. Due to time constraints we opted for grid search with following settings:
    \begin{table}[h]
        \centering
        \begin{tabular}{|l|l|l|}
            \hline
            Hyperparameter & Values & Best \\
            \hline
            LSTM Hidden Units & $2^{\{0, 1, 2, 3, 5, 7, 9\}}$ & 512 \\
            \hline
            LSTM Drop. & 0.0, 0.2, 0.4 & 0.0 \\
            \hline
            LSTM Recurr. Drop. & 0.0, 0.2, 0.4 & 0.0 \\
            \hline
            learning rate & 0.1, 0.001 & 0.001 \\
            \hline
        \end{tabular}
        \caption{Hyperparamter Search for POS experiment. Values investigated are in the middle column, while optimal combination of hyperparameters is in the right one.}
        \label{tab:pos-hypersearch}
    \end{table}

\begin{figure}[h]
        \centering
    	\includegraphics[scale=0.3]{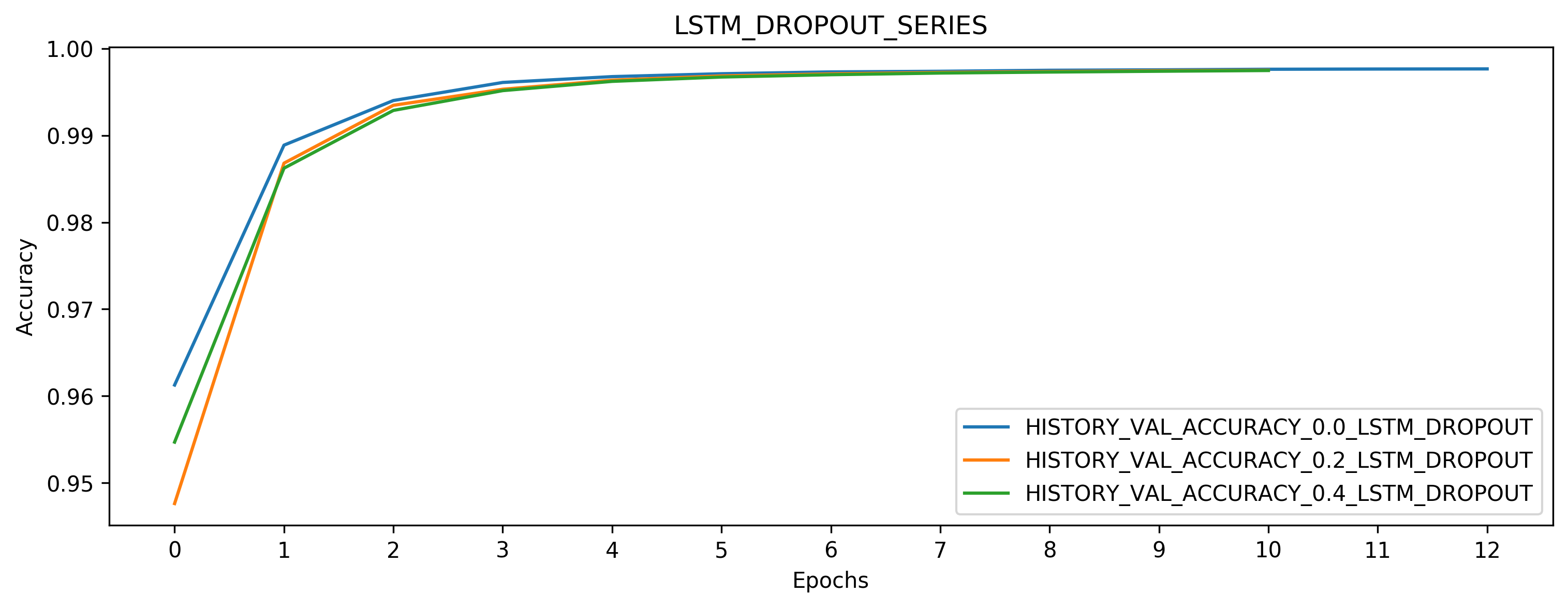}
        \caption{Performance of various settings for different dropout values in LSTM layer.}
        \label{fig:pos-hyperdrop}
    \end{figure}

    One of the surprises of optimal setting is that both of dropout's are zero. However, as seen on Fig. \ref{fig:pos-hyperdrop}, we can see that all settings converge to roughly the same performance on validation set, with $0.0$ having the fastest convergence.

    Sizes of Time Distributed Dense layer, and CRF layer are the same, and equal to number of classes given dataset. Additionally size of ELMo encoding (and hence size of input to BiLSTM layer) is 1024.

    Due to time constraints we also used early stopping. Training was stopped either after 40 epochs or if validation accuracy improvement was smaller than 0.001 over 4 epochs.

    Lastly, for simplicity of this paper, we are not discussing impact of all hyperparameters here. However, they are discussed in depth, with plots in the \href{https://github.com/andreaskuster/uw-nlp/tree/master/pos}{experiment on the github repository}.

    \subsubsection{Results}
    We tested the optimal model on 5 different setups, as described in list of flavors in Sec. \ref{sec:datasets}. The dataset used in this case was PTB, with splits as described before.

    \begin{table}[h]
        \centering
        \begin{tabular}{|l|c|c|c|}
            \hline
            Exp. & Test (C) & Test (U) & Avg \\
            \hline
            C        & 97.30 & 88.29 & 92.78 \\
            U        & 96.51 & 96.51 & 96.51 \\
            C + U    & 97.51 & 96.59 & \textbf{97.05} \\
            C + U 50 & 97.12 & 96.19 & \textbf{96.66} \\
            TT       & 95.04 & 95.04 & 95.04 \\
            TA       & 96.61 & 96.61 & 96.61 \\
            \hline
        \end{tabular}
        \caption{Accuracies for the POS task, in our setup. Averages of two best performing flavors are emboldened for readability.}
        \label{tab:pos-results-our}
    \end{table}

    \begin{table}[h]
        \centering
        \begin{tabular}{|l|c|c|}
            \hline
            Exp. & Our & Original \\
            \hline
            C        & 92.78 & 93.26 \\
            U        & 96.51 & 97.45 \\
            C + U    & \textbf{97.05} & \textbf{97.57} \\
            C + U 50 & \textbf{96.66} & \textbf{97.61} \\
            TT       & 95.04 & 95.21 \\
            TA       & 96.61 & 97.38 \\
            \hline
        \end{tabular}
        \caption{Comparison of average accuracies for the POS task. Results on the left are using our code, while ones of the left are reported in the original paper. For both sources 2 best results are emboldened.}
        \label{tab:pos-results-comp}
    \end{table}

    We can see that we agree with the hypothesis that mixing cased and uncased provides the best performance on the Penn Tree Bank as can be seen in Table \ref{tab:pos-results-our}. This confirms the hypothesis.

    However, as in case of Truecasing, we conclude that in terms of absolute performance we did get different results as reference paper. This can be seen in Table \ref{tab:pos-results-comp}, where there is about 1\% difference in accuracy between our implementation and the original one. In addition to this the gain due to mixing cased and uncased data is sometimes smaller than the difference between implementations, such as in case of Uncased flavor, where reference has higher accuracy than our C + U flavor.

    Overall, \textbf{we confirm the hypothesis}, but with a grain of salt due to lack of transparency in implementation in the original paper.

\subsection{Named Entity Recognition}
\label{sec:exp-ner}

    \subsubsection{Model description}
    Since the model wasn't described well in the original paper we used a fork of \href{https://github.com/ZhixiuYe/NER-pytorch}{NER-pytorch} which implements ~\cite{DBLP:journals/corr/LampleBSKD16}. This implementation has a similar performance to that of the original paper.

    This model uses pre-trained GloVe embeddings concatenated with character embeddings trained on the training data using a bi-directional LSTM. The character embeddings length is one of the hyper-parameters for the model.

    Similar to the pos model, it uses a single BiLSTM followed by a highway layer and and output CRF layer.

    \subsubsection{Hyperparameters}
    The character embedding size and number of hidden units in LSTM served as hyper-parameters for the model. We evaluated the model on the validation set for character embeddings size 25 and 40 along with LSTM hidden layer size of 200, 300 and 400 and found that (25,200) performed slightly better. LSTM Dropout and learning rate could have been other hyper-parameters we could've optimized but we maintained them at  0.5 and 0.015 due to time constraints.

    \subsubsection{Results}
    We tested on the same 5 variants as in case of POS experiment. Dataset used was CoNLL2003, similarly to the original paper. Table \ref{tab:ner-results-our} shows results of these variants using our code. We can see that, mixing cased and uncased data is again a very good data pre-processing step, which provides respectively first and third highest F1 score.

    \begin{table}[h]
        \centering
        \begin{tabular}{|l|c|c|c|}
            \hline
            Exp. & Test (C) & Test (U) & Avg \\
            \hline
            C        & 90.63 & 81.47 & 86.05 \\
            U        & 89.72 & 89.72 & \textbf{89.72} \\
            C + U    & 90.10 & 88.65 & 89.38 \\
            C + U 50 & 90.84 & 89.54 & \textbf{90.19} \\
            TT       & 80.89 & 80.89 & 80.89 \\
            TA       & 88.43 & 88.43 & 88.43 \\
            \hline
        \end{tabular}
        \caption{F1 scores for NER task, in our setup. Averages of two best performing variants are emboldened.}
        \label{tab:ner-results-our}
    \end{table}

    Table \ref{tab:ner-results-comp} compares scores from our implementation to ones from \cite{ner-and-pos-original}. We can see there is a 1-2\% gap in absolute results between our implementation and the original. This can happen due to minor differences in the implementations.

    \begin{table}[h]
        \centering
        \begin{tabular}{|l|c|c|}
            \hline
            Exp. & Our & Original \\
            \hline
            C        & 86.05          & 63.46 \\
            U        & \textbf{89.72} & 89.32 \\
            C + U    & 89.38          & \textbf{90.49} \\
            C + U 50 & \textbf{90.19} & \textbf{90.37} \\
            TT       & 80.89          & 82.93 \\
            TA       & 88.43          & 90.25 \\
            \hline
        \end{tabular}
        \caption{Comparison of average F1 scores for the NER task. Results on the left are using our code, while ones of the left are reported in the original paper. For both sources 2 best results are emboldened.}
        \label{tab:ner-results-comp}
    \end{table}

    However, the most interesting difference in the cased variant, where there is a above 30\% gap between our and the original implementation. After closer investigation we discovered that the reason for it is huge different in its performance on uncased data (81.47 in our implementation vs 34.46 in original one). We do not have a firm intuition on why this is happening. However, it might be the case that models trained on cased dataset are highly unstable when tested on uncased data.

    Overall, however we can see that relative performance of results is similar, and mixing cased and uncased data provides the best performance with our implementation. Because of this we believe \textbf{our results support second hypothesis of the paper}.

    Another experiment that \cite{ner-and-pos-original} performed is testing how well their model from NER task transfers to other datasets. In particular they used the Twitter dataset, since it has very different properties than CoNLL2003.

    Table \ref{tab:ner-twitter-comp} compares performance of our implementation to the original one when transferred to the Twitter dataset. We see that our implementation performs much worse than the original one (by 30\% or more in each case).

    \begin{table}[h]
        \centering
        \begin{tabular}{|l|c|c|}
            \hline
            Exp. & Our & Original \\
            \hline
            C        & 33.24 & 58.63 \\
            U        & 14.54 & 53.13 \\
            C + U    & 31.31 & 66.14 \\
            C + U 50 & 32.94 & 64.69 \\
            TT       & 23.45 & 58.22 \\
            TA       & 29.19 & 62.66 \\
            \hline
        \end{tabular}
        \caption{Comparison of average F1 scores for the NER task, when transferred to the Twitter dataset. Results on the left are using our code, while ones of the left are reported in the original paper. Since, it is a direct comparison we do not embold any results.}
        \label{tab:ner-twitter-comp}
    \end{table}

    Comparison here is not relative, but direct, on absolute values, due to the fact that we are testing our performance on a new dataset. This means, that our model did much worse than the original one. This is very counterintuitive, when we consider the fact that in the original dataset, our cased experiment generalized much better. Our intuition is that there was a dropout used in the original implementation, which was undocumented in the paper.

    Overall, we \textbf{cannot support the thrid hypothesis from original paper}.

\section{Experiments beyond the original paper}

    \subsection{Datasets}
    In addition to the datasets used in original paper, we additionally used:
    \begin{itemize}
        \item Brown - From \href{http://korpus.uib.no/icame/brown/bcm.html}{Brown University}. We used a loader from \href{https://www.nltk.org/}{nltk}.
        \item CoNLL2000 - From \cite{conll2000}. We used a loader from \href{https://www.nltk.org/}{nltk}.
        \item PTB Reduced (PTB R) - Same as Penn Tree Bank described in Section \ref{sec:datasets}, but:
        \begin{itemize}
            \item Train consists of sections 0-4
            \item Validation consists of sections 5-6
            \item Test consists of sections 7-8
        \end{itemize}
    \end{itemize}

    \subsection{Part of Speech Tagging}
    The aim of the additional experiments is to find out if the hypothesis from the paper is more generally applicable. We run the same experiments on LSTM models with different word embeddings, with or without the CRF layer, and on different datasets.

    \subsubsection{Hyperparameters}
    Due to results in Section \ref{sec:exp-pos-hyper} the same model was used as in the original experiment.

    One important note for encodings is that they change number of inputs to BiLSTM layer, thus slightly modifying its size. In particular both GloVe and word2vec have size of 300, which is much smaller than 1024 in ELMo.

    \subsubsection{Results}
    Since there were 3 experiments roughly separate experiments let us present them separately.

    \begin{table}[h]
        \centering
        \begin{tabular}{|l|c|c|c|}
            \hline
            Exp.     & word2vec       & GloVe          & ELMo \\
            \hline
            C        & 83.71          & 91.01          & 92.78 \\
            U        & 80.97          & 94.67          & 96.51 \\
            C + U    & \textbf{86.15} & \textbf{96.36} & \textbf{97.05} \\
            C + U 50 & 85.01          & \textbf{95.35} & \textbf{96.66} \\
            TT       & 85.74          & 93.82          & 95.04 \\
            TA       & \textbf{86.64} & 95.20          & 96.61 \\
            \hline
        \end{tabular}
        \caption{Average accuracies for different encodings. Note that ELMo encodings are the exact copy of results from \ref{tab:pos-results-our}. Similarly to other tables 2 best scores are emboldened for readability.}
        \label{tab:epos-encodings}
    \end{table}

    Table \ref{tab:epos-encodings} shows results for various encodings. We see that we consistently mixed dataset appears to be in the top 2 results. We can also see that encoding has a strong effect on absolute performance of the model, especially in case of word2vec which has a 10\% drop in accuracy relative to GloVe and ELMo. This is an expected behavior as ELMo is known to outperform both word2vec and GloVe in many cases.

    There is an interesting notion that both of Truecasing flavors tend to perform better than Half Mixed one. However, due to significant drop in word2vec with respect to other encodings we believe that it rather an artifact of the encoding itself rather than a theme. Additionally, performance of all three scenarios is rather close further pointing to encoding specific problem.

    Overall, we believe that the first additional hypothesis is \textbf{supported by the above results}.

    \begin{table}[h]
        \centering
        \begin{tabular}{|l|c|c|c|}
            \hline
            Exp. & PTB R & Brown & CoNLL2000 \\
            \hline
            C        & 92.36          & 89.50          & 92.86\\
            U        & 95.48          & 92.91          & 96.83 \\
            C + U    & \textbf{96.22} & \textbf{96.47} & \textbf{99.23} \\
            C + U 50 & \textbf{95.71} & \textbf{93.92} & \textbf{97.16} \\
            TT       & 94.62          & 92.11          & 95.40 \\
            TA       & 95.35          & 92.62          & 96.79 \\
            \hline
        \end{tabular}
        \caption{Average accuracies for different datasets.}
        \label{tab:epos-results-datasets}
    \end{table}

    Table \ref{tab:epos-results-datasets} shows results for different dataset. Here results are clear, mixing cased and uncased datasets provides the best result regardless of a dataset. In fact for Brwon CoNLL2000 gaps between this technique and others is even larger than on the original PTB dataset. Hence, these results \textbf{strongly support our second additional hypothesis}.

    \begin{table}[h]
        \centering
        \begin{tabular}{|l|c|c|}
            \hline
            Exp. & No CRF & CRF \\
            \hline
            C        & 92.88 & 92.78 \\
            U        & 96.52 & 96.51 \\
            C + U    & 97.02 & 97.05 \\
            C + U 50 & 96.83 & 96.66 \\
            TT       & 94.78 & 95.04 \\
            TA       & 96.56 & 96.61 \\
            \hline
        \end{tabular}
        \caption{Average accuracies for the original model with CRF and without CRF.}
        \label{tab:epos-results-crf}
    \end{table}

    Table \ref{tab:epos-results-crf} compares performance of the original model with or without the last layer. Considering that most of results are within a few hundredths between two models, with TT being 0.26\% away from the original implementation, we conclude that CRF can be eliminated from the model without major differences in performance. This is very beneficial on the implementation and runtime side, since CRF is a non-standard library, making model are complicated, and does not have multi-gpu support, causing it to run much longer. These results, \textbf{strongly support our third additional hypothesis}, and we recommend potential future users of BiLSTM to not include CRF layer (at least for POS task).

\section{Computational requirements}
Due to relative lack of overlap between three sections each of us used different computational resources:
\begin{itemize}
    \item Truecasing:
    \begin{itemize}
        \item GPU: NVIDIA RTX 2080 Super
        \item CPU: AMD Ryzen 7 3700x
        \item Runtime: 10h (Original Paper)
    \end{itemize}
    \item POS:
    \begin{itemize}
        \item GPU: 2x NVIDIA V100
        \item CPU: AMD EPYC 7501
        \item Runtime: 100h (Original Paper + Additional Experiments)
    \end{itemize}
    \item NER:
    \begin{itemize}
        \item GPU: NVIDIA GTX 1080
        \item CPU: Intel Core i7-6850K
        \item Runtime: 26h (Original Paper + Twitter Evaluation)
    \end{itemize}
\end{itemize}

This is required for every report. Include every item listed in the syllabus, plus any other relevant statistics (e.g. if you have multiple model sizes, report info for each).
Include the information listed in the syllabus, including the \textbf{total} number of GPU hours used for all experiments, and the number of GPU hours for \textbf{each} experiment.

\section{Discussion and recommendations}
\cite{ner-and-pos-original} is mostly reproducible, with the three out of four original hypothesis reproduced in our report. In particular all hypotheses which were not involving transferring task to a new dataset were successful.

We believe that transferring NER to Twitter is also a feasible claim, but involving a lot of additional tunning to the model, which should be described in the original paper. This is because of our additional experiments in POS, which shown that at least in that task, the same type of model is transferrable to other datasets.

Most of the elements of this paper worked surprisingly well when reproduced, and implementation was not a significant struggle, due to libraries such as PyTorch and Keras. Our main concern is that if we were not provided access to datasets by TAs, it would be a much harder task, since two datasets used in the original paper were not available publicly. Our suggestion to future researches is to use publicly available datasets, if possible.

Overall, we would suggest future researchers to use a mixed cased and uncased data in their work, since it provide a little development time overhead and no additional runtime.

\bibliographystyle{acl_natbib}
\bibliography{emnlp-ijcnlp-2019}

\end{document}